\pdfoutput=1

\documentclass[11pt]{article}

\usepackage[final]{acl}

\usepackage{times}
\usepackage{latexsym}

\usepackage[T1]{fontenc}

\usepackage[utf8]{inputenc}

\usepackage{microtype}

\usepackage{multirow}
\usepackage{graphicx}
\usepackage{color}
\usepackage{hyperref}
\usepackage{microtype}
\usepackage{lscape}
\usepackage{longtable}
\usepackage{booktabs} 
\usepackage{tabularx} 
\usepackage{lipsum} 
\usepackage{times}
\usepackage{siunitx}
\usepackage{makecell}
\usepackage{latexsym}
\usepackage{amsmath}
\usepackage{mdframed,lipsum}
\usepackage[most]{tcolorbox}

\title{\texorpdfstring{S\textsc{ci}P\textsc{rompt}: Knowledge-augmented Prompting for Fine-grained Categorization of Scientific Topics}{S\textsc{ci}P\textsc{rompt}: Knowledge-Augmented Prompting for Fine-Grained Categorization of Scientific Topics}}

\author{Zhiwen You$^1$, Kanyao Han$^1$, Haotian Zhu$^2$,
\textbf{Bertram Lud\"ascher$^1$, Jana Diesner$^{1,3}$} \\
$^1$ University of Illinois Urbana-Champaign \\
$^2$ University of Washington \\
$^3$ Technical University of Munich \\
\texttt{\{zhiweny2, kanyaoh2, ludaesch\}@illinois.edu} \\
\texttt{haz060}\texttt{@uw.edu} \quad \texttt{jana.diesner}\texttt{@tum.de}}

\begin{document}
\maketitle
\begin{abstract}
Prompt-based fine-tuning has become an essential method for eliciting information encoded in pre-trained language models for a variety of tasks, including text classification. For multi-class classification tasks, prompt-based fine-tuning under low-resource scenarios has resulted in performance levels comparable to those of fully fine-tuning methods. Previous studies have used crafted prompt templates and verbalizers, mapping from the label terms space to the class space, to solve the classification problem as a masked language modeling task. However, cross-domain and fine-grained prompt-based fine-tuning with an automatically enriched verbalizer remains unexplored, mainly due to the difficulty and costs of manually selecting domain label terms for the verbalizer, which requires humans with domain expertise. To address this challenge, we introduce S\textsc{ci}P\textsc{rompt}, a framework designed to automatically retrieve scientific topic-related terms for low-resource text classification tasks. To this end, we select semantically correlated and domain-specific label terms within the context of scientific literature for verbalizer augmentation. Furthermore, we propose a new verbalization strategy that uses correlation scores as additional weights to enhance the prediction performance of the language model during model tuning. Our method outperforms state-of-the-art, prompt-based fine-tuning methods on scientific text classification tasks under few and zero-shot settings, especially in classifying fine-grained and emerging scientific topics\footnote{Our code is available at \url{https://github.com/zhiwenyou103/SciPrompt}.}. 
\end{abstract}


\begin{figure*}[ht]
  \centering
  \includegraphics[width=0.88\textwidth]{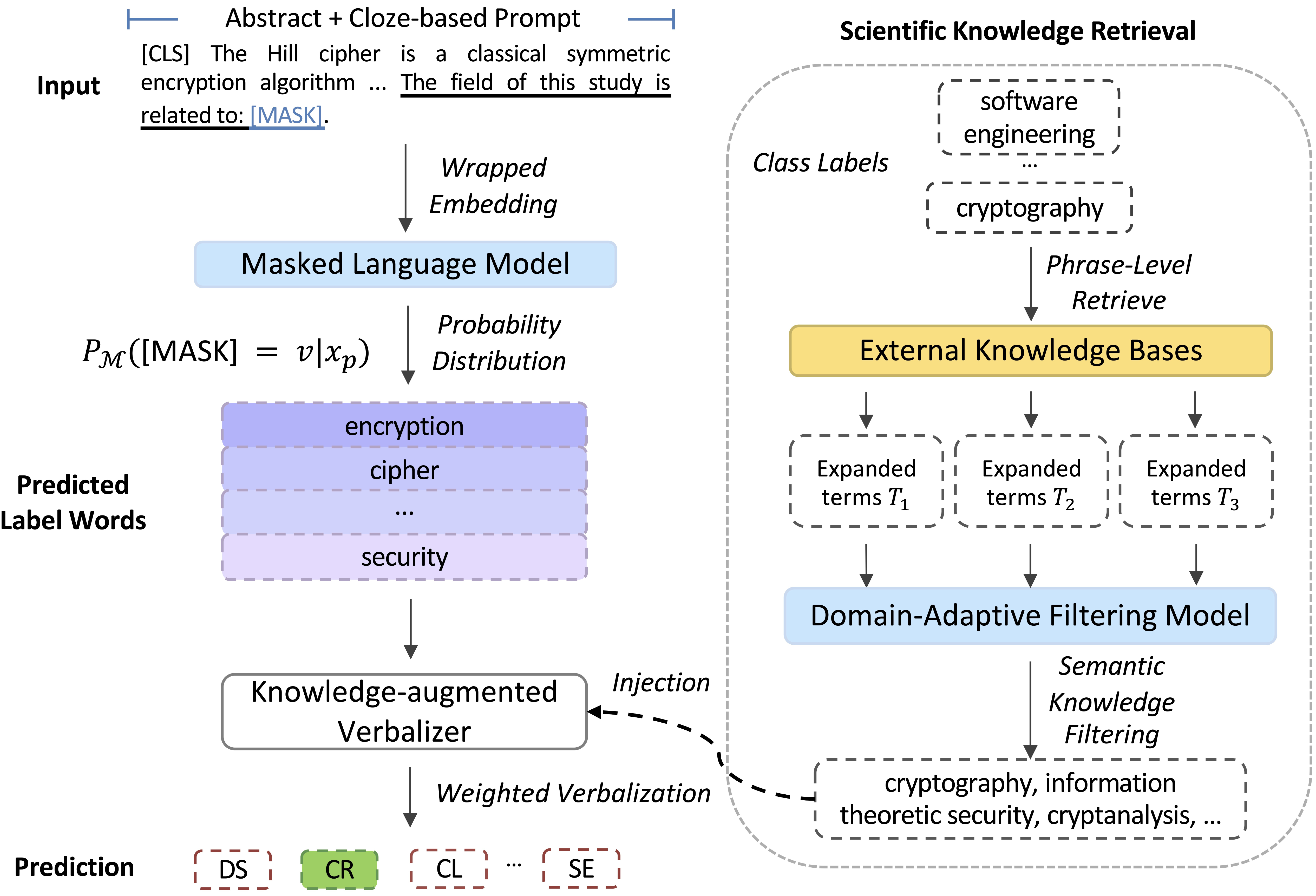}
  \caption{Overall framework of S\textsc{ci}P\textsc{rompt}. The left side shows the overall process of masked language modeling for performing the text classification task. The right side shows our proposed knowledge retrieval and domain-adaptive filtering phase (\S\ref{sec:method}). The prediction results, such as CR and SE, correspond to the class labels for Cryptography and Software Engineering, respectively, and are used for scientific knowledge retrieval.}
  \label{fig:framework}
\end{figure*}

\section{Introduction}
Scientific text classification tasks involve categorizing scientific abstracts into specific disciplines or topics. Recent studies leverage prompt-based fine-tuning method \citep{ding2022prompt, gu2022ppt, schick2020exploiting, liu2023pre}, transferring the text classification problem as a masked language modeling task. Masked Language Models (MLMs) are developed by extensively training on large text corpora with a percentage of the input tokens being randomly replaced with a \texttt{[MASK]} token. Traditional fine-tuning, which requires additional training on labeled domain- or task-specific data \cite{ovadia2023fine}, may not be suitable in limited data scenarios, such as few and zero-shot settings. Prompt-based fine-tuning has emerged as an effective alternative. This approach uses a prompt to guide the MLM in generating a specific token through masking a \texttt{[MASK]} token in the prompt template, addressing the text classification tasks \citep{schick2020exploiting, hu2021knowledgeable, chen2022adaprompt, gao-etal-2021-making} under low-resource conditions (i.e., few and zero-shot settings) through a verbalizer. As defined by \citet{schick2020exploiting}, the verbalizer refers to the mapping from label words (e.g., ``cryptanalysis'') to the corresponding class (e.g., ``Cryptography''), serving as a projection function between the vocabulary and the class label space. However, in the context of classifying scientific literature, the complexity of scientific language and scarcity of fine-grained (i.e., a wide range of scientific fields that are labeled with sub-categories) or emerging topics make it hard to automatically classify cross-domain scholarly articles with limited training samples and manually created verbalizers \cite{schick2020exploiting}. 

The goal of this paper is to address the challenge of multi-class classification in low-resource settings, specifically focusing on classifying scientific abstracts into different domains with only a limited number of labeled examples. We introduce a prompt-based fine-tuning approach enriched with domain knowledge as a new strategy for retrieving domain-adaptive label terms (i.e., scientific terms in various fields) without manual intervention. We enhance our approach for low-resource scenarios by retrieving scientific phrases from external knowledge bases (KBs) to expand label terms of the verbalizer \cite{hu2021knowledgeable} from the token-level to term phrases. We fine-tune Natural Language Inference (NLI) models for semantic similarity search between retrieved label terms and class labels to select domain-related scientific phrases. Our method differs from previous studies \cite{hu2021knowledgeable, ding2022openprompt}, which rely on word frequency filtering and are limited to single-token verbalizer projection for text classification. Given the complexity of scientific terminology (see Appendix~\ref{sec:appendixC} for more details), we refine the traditional verbalization approach \cite{ding2022prompt} by integrating scientific terms through deploying a weight-aware label term mapping function. This approach improves the projection performance from MLM's predictions to probabilities of a specific class compared with prior studies \cite{hu2021knowledgeable, gao2021making, chen2022decoupling}.

Our approach consists of three stages: 1) retrieval of scientific terms, 2) label term filtering, and 3) prediction of scientific topics. Initially, we use a cloze-style prompt and an input scientific abstract to guide the MLM to generate label words to fill the \verb|[MASK]| token (Figure~\ref{fig:framework}). Then, we use each class label as a query to retrieve class-related domain phrases (also denote as ``label terms'') from external KBs. To filter the potentially irrelevant terms gathered in the retrieval stage, we fine-tune both bi-encoder and cross-encoder models using the SciNLI dataset \cite{sadat2022scinli}, enabling the selection of the most relevant domain phrases. Finally, with the selected sets of knowledge-enriched scientific terms, we incorporate these label terms into the verbalizer to convert the MLM's prediction into a specific class through a semantic score-weighted average loss, enhancing the precision of the probability projections for the augmented verbalizer. Our method extends beyond token-to-token verbalization by encompassing token-to-phrase verbalization that enriches the semantic meaning of scientific domain vocabulary. This broader scope allows for an advanced interpretation of scientific language and classifying emerging topics under weak supervision.

In summary, our contributions are the presentation of: 
\begin{itemize}
    \item A domain-adaptive prompt-based fine-tuning framework, named S\textsc{ci}P\textsc{rompt}, for fine-grained and low-resource scientific text classification tasks.
    \item A new knowledge retrieval and filtering strategy for automatically enriching the verbalizer with domain knowledge.
    \item A weighted verbalization approach tailored for mapping filtered scientific label terms from model predictions to specific classes.
    \item Evaluation via experiments on four scientific datasets show that S\textsc{ci}P\textsc{rompt} largely outperforms most state-of-the-art methods in few and zero-shot settings.
\end{itemize}

\section{Related Work} \label{sec:relatedwork}
\subsection{Knowledge-Powered Prompting for Text Classification}
A Pattern-Exploiting Training (PET) framework \cite{schick-schutze-2021-exploiting, schick-schutze-2021-just}, which initially investigated how cloze-based prompt templates can guide language models to tackle classification tasks \cite{han2022ptr, ding2022openprompt, min-etal-2022-noisy, wang2022automatic, zhang2022prompt, wang2022towards}, has inspired research on incorporation more diverse label words into the verbalizer. Specifically, \citet{hu2021knowledgeable} added external knowledge to the verbalizing process to help an MLM predict masked tokens more accurately. AdaPrompt \cite{chen2022adaprompt} applied a different knowledge injection method that leveraged task and prompt characteristics to retrieve external knowledge for continuous pre-training of MLMs adaptively. 
However, classifying scientific literature presents challenges that previous methods have not addressed, including projecting phrase-level label terms in the verbalization process. Other challenges, to which a broad range of solutions have been developed, include handling complex semantic structures in a wide range of scientific topics \cite{eykens2021fine, khadhraoui2022survey} and the scarcity or imbalance of labeled data across multiple disciplines \cite{cunha2021cost}. 


\subsection{Label Terms Refinement}
Prior research on prompt-based fine-tuning has used the verbalizer module to map MLM's predictions to specific classes. \citet{schick-schutze-2021-exploiting} introduced an automatic verbalizer search that identifies suitable label words from training data and language models to enrich the verbalizer. This approach has been further explored in different studies to improve the classification performance \citep{gao-etal-2021-making, shin-etal-2020-autoprompt, liu2021gpt}, although these methods typically need extensive training data, making them less suitable for low-resource scenarios. To address these challenges, one can manually expanding the verbalizer with more label words \citep{shin-etal-2020-autoprompt}, which has limitations when classifying fine-grained and domain-related categories that need expert knowledge. Recently, external KBs have been used to enrich the verbalizer by sourcing class-related label words \citep{hu2021knowledgeable, chen2022adaprompt}.

\section{Methodology} \label{sec:method}
Our framework of S\textsc{ci}P\textsc{rompt} uses a two-stage approach for scientific text classification: (1) masked language modeling and (2) domain knowledge retrieval and filtering. 

\subsection{Cloze-Style Masked Language Modeling} \label{sec:3.1}
MLMs $\mathcal{M}$ (e.g., SciBERT \cite{beltagy2019scibert}) are created by randomly masking tokens in the training text and training the model to predict the masked tokens. Similarly, prompt-based fine-tuning typically leverages a cloze- or prefix-based prompt template, reformulating the input into a masked language modeling task. This strategy enables $\mathcal{M}$ to predict the masked token, facilitating the execution of downstream tasks based on $\mathcal{M}$ outputs. Building upon \citet{hu2021knowledgeable}, our framework employs a few-shot prompt-based fine-tuning strategy that conceptualizes scientific text classification as an \textit{N}-way \textit{K}-shot task, where \textit{N} indicates the number of classes and \textit{K} is the number of labeled examples per class.

We provide a limited number of labeled examples for each class to tune $\mathcal{M}$. We construct a training $\mathcal{D}_{train}$ and validation set $\mathcal{D}_{val}$ following previous studies \cite{gao-etal-2021-making, perez2021true, wang2022automatic, hu2021knowledgeable} with $n$ examples per class. For the few-shot setting, given a cloze-based prompt template $\mathcal{P}_t$ and an input abstract $a_n$, where $a_n \in \mathcal{D}_{train}$, $\mathcal{M}$ predicts the label word $l$ to fill into the \verb|[MASK]| position in the prompt template. After that, the verbalizer function $f_v$ maps the predicted label word $l$ onto pre-defined label term set $\mathcal{L}$ to classify it into a class, i.e., $\mathcal{L} \rightarrow \mathcal{Y}$. We use a cross-entropy loss \cite{gao-etal-2021-making} to update the parameters of $\mathcal{M}$ through verbalization outputs. For instance, the prompt is designed as ``\verb|[Abstract]|. The field of this article is related to: \verb|[MASK]|''. $\mathcal{M}$ will predict suitable label word $l$ to fill into the \verb|[MASK]|. Then, $f_v$  calculates the probability of classifying $l$ into a topic $y_i$, where $y_i \in \mathcal{Y}$:
\begin{equation}
    \begin{gathered}
P(\mathcal\!y_i\!\mid\!a_n)\!=f_v({P(\verb|[MASK]|\!=\!\mathcal{M}(l)\!\mid\!a_n\!)}), \label{eq1}
    \end{gathered}
\end{equation}
where $l \in \mathcal{L}$. In the zero-shot setting, given $\mathcal{M}$ can directly generate a label word to fill into \verb|[MASK]|, we use the output of $\mathcal{M}$ as the final label word and send the output into the verbalizing function to calculate class probabilities without tuning loss updates.

\subsection{Scientific Knowledge Retrieval} \label{sec:3.2}
Predicting masked tokens using an MLM involves generating a range of potential label words, each with varying probabilities of matching a specific class. Enhancing the verbalizer with a more extensive set of label terms has been proven to improve the accuracy of word-to-class mapping \cite{{hu2021knowledgeable, chen2022adaprompt, wang2022automatic, shin-etal-2020-autoprompt}}. To implement this approach, we use two external KBs, \texttt{Related Words}\footnote{\url{https://relatedwords.org}} and \texttt{Reverse Dictionary}\footnote{\url{https://reversedictionary.org}} for scientific knowledge retrieval. \texttt{Related Words} identifies relevant terms using vector similarity and resources like word embeddings and ConceptNet. \texttt{Reverse Dictionary}, which acts as a word search engine, finds terms based on definitions or phrases. \texttt{Reverse Dictionary} is particularly useful in phrase-level retrieval, where straightforward labels from \texttt{Related Words} may not suffice given a domain-specific phrase (e.g., Networking and Internet Architecture). We set class labels $\mathnormal{C} = \{y_1, y_2, ..., y_n\} $ as queries to retrieve from \texttt{Related Words} $\mathcal{G}_{RW}$.

When $\mathcal{G}_{RW}$ fails to produce terms with similarity scores above zero, we use \texttt{Reverse Dictionary}, denoted as $\mathcal{G}_{RD}$, for additional phrase retrieval. 
Each retrieved term is assigned a single relevance score. Initially, we adopted the same threshold (i.e., threshold = 0) as KPT \cite{hu2021knowledgeable} for term retrieval based on topic names. Subsequently, we impose two additional thresholds for further selection of retrieved terms (\S\ref{sec:3.3}). Utilizing these KBs enables the compilation of knowledge-enhanced term sets $\mathcal{T}_i={t_1, t_2, ..., t_m}$ for each dataset, where $i \in n$ and $t$ represents the retrieved label terms. Note that the number of terms $m$ may vary for each class.

\subsection{Domain Adaptive Model Tuning} \label{sec:3.3}
To effectively identify the most relevant label words for each class from a set of initial raw terms, it is crucial to use a model tailored or adaptable to specific fields. Drawing from \citet{chen2022adaprompt}, who employed a pre-trained NLI model to filter label words produced by an MLM, we present a method that enhances the accuracy of selecting label terms related to specific topics by integrating domain knowledge. We apply a newly introduced scientific NLI dataset $\mathcal{D}_{SciNLI}$ \cite{sadat2022scinli}, consisting of labeled sentence pairs $(s_i, s_j)$ from scholarly articles in the fields of NLP and computational linguistics. This dataset serves to fine-tune both cross-encoder $\mathcal{M}{ce}$ and bi-encoder $\mathcal{M}_{be}$ NLI models\footnote{\url{https://www.sbert.net/examples/applications/cross-encoder/}}, where $\mathcal{M}_{be}$ produces for a given sentence a sentence embedding and $\mathcal{M}_{ce}$ passes a sentence pair to the encoder to produce an output value between 0 and 1 indicating the similarity of the input sentence pair \cite{reimers2019sentence}. The training labels are defined as {``\textit{Entailment}'' or ``\textit{Contradiction}'',} thus framing the model fine-tuning as a binary classification task:
\begin{equation*}
    \begin{aligned}
    \mathcal{M'}(s_i, s_j) = \left\{
        \begin{array}{rcl}
        >0 & & {\text{if } s_i \text{ entails } s_j}\\
        <0 & & {\text{if } s_i \text{ contradicts } s_j}\\
        \end{array}, \right. 
    \end{aligned}
\end{equation*}
where $\mathcal{M'}$ denotes either $\mathcal{M}_{ce}$ or $\mathcal{M}_{be}$.

\subsection{Semantic Knowledge Filtering} \label{sec:3.4}
We merge each retrieved scientific label term with a standard prompt (see Appendix~\ref{sec:appendixD}), encode prompts using the fine-tuned $\mathcal{M}_{be}$, and use these encoded embeddings as queries for sentence-level semantic searches to select topic-related label terms and calculate semantic similarity scores $w_l$ for each label term. We apply SentenceTransformers\footnote{\url{https://www.sbert.net/index.html}} to conduct the cosine similarity search using $\mathcal{M}_{be}$ within each retrieved label term set. Then, we use $\mathcal{M}_{ce}$ to re-rank these label terms for every prompt pair of each topic, selecting relevant sentences based on predefined thresholds ($\mu_{be} = 0.5$, $\mu_{ce} = 0.1$). As these scores also help predict label words, we apply this method in few and zero-shot scenarios (for more details, see Appendix~\ref{threshold}).

Following KPT \cite{hu2021knowledgeable}, we also apply a label term calibration approach with a full training set to directly remove irrelevant label terms in the verbalizer that are less likely to be predicted by $\mathcal{M}$. The retrieved label terms for each class with lower probabilities (i.e., less than 0.5) produced by $\mathcal{M}$ are removed. The probability of $t$ is:
\begin{equation}\label{eq2}
    \hat{P}_{\mathcal{M}}(\texttt{[MASK]}=t|a_n)\propto\frac{P_{\mathcal{M}}(\texttt{[MASK]}=t|a_n)}{prior(p_t)},
\end{equation}
where $prior(p_t)$ is the prior probability of the label term $t$ produced by $\mathcal{M}$ using the training set.

\subsection{Weighted Verbalizer Transformation} \label{sec:3.5}
Given that retrieved label terms may be tokenized into multiple tokens, we adopt a ``mean'' method to average the tokens of a label term \cite{ding2022openprompt}, considering all parts of a term as significant. 

Adopting the verbalizer structure from \citet{ding2022openprompt}, we introduce a verbalization approach that maps $\mathcal{M}$'s output to specific classes $y_i$, using predefined semantic scores $w_l$ as weights for each label term. This method aims to enhance the accuracy of classifying $\mathcal{M}$'s predictions $l$ into topic $y_i$:
\begin{equation}
    \begin{aligned}
P(y_i|a_n)&=\mathop{\arg\max}\limits_{y_i \in \mathcal Y} s(v_{y_i}|h_{mask}, w_l) \\
&=\frac{\exp{(v_{y_i}\cdot h_{mask}\cdot w_l})}{\sum_{y \in \mathcal{Y}} \exp{(v_{y}\cdot h_{mask}\cdot w_l})},\label{eq4}
    \end{aligned}
\end{equation}
where the objective function $s(v_{y_i}|h_{mask}, w_l)$ calculates $\mathcal{M}$'s probability for the output $v_{y_i}$ of the \verb|[MASK]| token, with $v_{y_i}$ as the label term embeddings, and $h_{mask}$ as the hidden states at the \verb|[MASK]| position. This objective function can be optimized through the cross-entropy loss as denoted in Equation~(\ref{eq1}).

\subsection{Vector-Based Verbalizer Mapping} \label{sec:3.6}
Incorporating the filtered label terms into the verbalizer is crucial for making accurate predictions and eliminating noise simultaneously. Moving beyond simple summing \cite{wang2022automatic} or weighted averaging \cite{hu2021knowledgeable} of label words, the Word-level Adversarial ReProgramming (WARP) model introduced in \cite{hambardzumyan-etal-2021-warp} uses vector representations for class mapping, which is distinct from conventional single word projection. We introduce a new method named S\textsc{ci}P\textsc{rompt}$_{\textit{Soft}}$ based on the uniqueness of our phrase-level verbalizer. Specifically, we refine the verbalization in S\textsc{ci}P\textsc{rompt}$_{\textit{Soft}}$ by drawing from the soft verbalizer concept introduced by WARP. In the experiments with $\text{S\textsc{ci}P\textsc{rompt}}_{\textit{Soft}}$, we aggregate all retrieved label terms per topic with semantic scores into a vector for topic probability prediction and optimize the aggregated vector during model tuning (detailed in Appendix~\ref{sec:appendixB}).

\section{Experiments}
We present the experimental settings of S\textsc{ci}P\textsc{rompt} across scientific classification datasets in few and zero-shot scenarios.

\subsection{Datasets} \label{sec:4.1}
We use three publicly available datasets in English for our experiments: SDPRA 2021 \cite{reddy2021overview}, arXiv \cite{meng2019weakly}, and S2ORC \cite{lo2019s2orc}.
SDPRA 2021 contains scientific articles from computer science across seven categories. arXiv \cite{meng2019weakly} includes abstracts sourced from the arXiv website\footnote{\url{https://arxiv.org/}} across 53 sub-categories, and S2ORC contains academic papers from across 19 disciplines. For the S2ORC data, we only select abstracts with a single discipline label through the Semantic Scholar Public API\footnote{\url{https://www.semanticscholar.org/product/api}}. 
The statistics and category examples of these datasets are shown in Table~\ref{tab:datasets-stat} and Appendix~\ref{sec:appendixC}.  


\begin{table*}[!htb]
\centering
\footnotesize
\scalebox{1.1}{
\begin{tabular}{c|cccc|c}
\midrule[1.5pt]
\textbf{Examples} & \textbf{Method} & \textbf{SDPRA 2021} & \textbf{{arXiv}} & \textbf{S2ORC} & \textbf{Avg.}\\
\midrule[1pt] 
\multirow{5}{*}{1} & $\text{Fine-tuning}_{\textit{SciBERT}}$ & 12.72 \begin{small} ${\pm~3.70}$ \end{small} & 2.03 \begin{small} ${\pm~0.21}$ \end{small}& 4.76 \begin{small} ${\pm~0.85}$ \end{small} & 6.50 \begin{small} ${\pm~1.59}$ \end{small}\\
  & $\text{Prompt-tuning}_{\textit{Manual}}$ & \textbf{71.68} \begin{small} ${\pm~4.73}$ \end{small} & 34.95 \begin{small} ${\pm~1.45}$ \end{small} & 40.88 \begin{small} ${\pm~1.92}$ \end{small} & 49.17 \begin{small} ${\pm~2.70}$ \end{small}\\
  & LM-BFF & 68.95 \begin{small} ${\pm~1.68}$ \end{small} & 35.07 \begin{small} ${\pm~1.31}$ \end{small} & 41.50 \begin{small} ${\pm~1.43}$ \end{small} & 48.51 \begin{small} ${\pm~1.47}$ \end{small}\\
  & KPT & 50.74 \begin{small} ${\pm~3.03}$ \end{small} & 32.18 \begin{small} ${\pm~1.08}$ \end{small} & 43.20 \begin{small} ${\pm~1.33}$ \end{small} & 42.04 \begin{small} ${\pm~1.81}$ \end{small}\\
  & $\text{S\textsc{ci}P\textsc{rompt}}$ & 64.42 \begin{small} ${\pm~3.64}$ \end{small} & \textbf{40.57} \begin{small} ${\pm~1.60}$ \end{small} & \textbf{47.92} \begin{small} ${\pm~1.67}$ \end{small} & \textbf{50.97} \begin{small} ${\pm~2.30}$ \end{small}\\
  & $\text{S\textsc{ci}P\textsc{rompt}}_{\textit{Soft}}$ & 62.65 \begin{small} ${\pm~4.94}$ \end{small} & 31.06 \begin{small} ${\pm~1.74}$ \end{small} & 29.94 \begin{small} ${\pm~1.94}$ \end{small} & 41.22 \begin{small} ${\pm~2.87}$ \end{small}\\ 
\midrule
\multirow{6}{*}{5} & $\text{Fine-tuning}_{\textit{SciBERT}}$ & 16.45 \begin{small} ${\pm~4.35}$ \end{small} & 2.36 \begin{small} ${\pm~0.55}$ \end{small} & 5.63 \begin{small} ${\pm~1.37}$ \end{small} & 8.15 \begin{small} ${\pm~2.09}$ \end{small}\\
  & $\text{Prompt-tuning}_{\textit{Manual}}$ & 83.46 \begin{small} ${\pm~1.41}$ \end{small} & 47.58 \begin{small} ${\pm~1.68}$ \end{small} & 49.53 \begin{small} ${\pm~0.88}$ \end{small} & 60.19 \begin{small} ${\pm~1.32}$ \end{small}\\
  & LM-BFF & 79.97 \begin{small} ${\pm~2.52}$ \end{small} & 50.11 \begin{small} ${\pm~0.88}$ \end{small} & 48.67 \begin{small} ${\pm~1.02}$ \end{small} & 59.58 \begin{small} ${\pm~1.47}$ \end{small}\\
  & RetroPrompt & 64.76 \begin{small} ${\pm~3.57}$ \end{small} & 31.37 \begin{small} ${\pm~0.72}$ \end{small} & 47.09 \begin{small} ${\pm~1.38}$ \end{small} & 47.74 \begin{small} ${\pm~1.89}$ \end{small}\\
  & KPT & 77.71 \begin{small} ${\pm~3.34}$ \end{small} & 53.68 \begin{small} ${\pm~1.69}$ \end{small} & 50.40 \begin{small} ${\pm~1.84}$ \end{small} & 60.60 \begin{small} ${\pm~2.29}$ \end{small}\\
  & $\text{S\textsc{ci}P\textsc{rompt}}$ & 81.81 \begin{small} ${\pm~3.34}$ \end{small} & 56.36 \begin{small} ${\pm~0.95}$ \end{small} & \textbf{52.12} \begin{small} ${\pm~1.59}$ \end{small} & \textbf{63.43} \begin{small} ${\pm~1.96}$ \end{small}\\
  & $\text{S\textsc{ci}P\textsc{rompt}}_{\textit{Soft}}$ & \textbf{83.70} \begin{small} ${\pm~2.86}$ \end{small} & \textbf{58.01} \begin{small} ${\pm~0.94}$ \end{small} & 47.44 \begin{small} ${\pm~1.60}$ \end{small} & 63.05 \begin{small} ${\pm~1.80}$ \end{small}\\
\midrule
\multirow{6}{*}{10} & $\text{Fine-tuning}_{\textit{SciBERT}}$ & 17.44 \begin{small} ${\pm~4.50}$ \end{small} & 3.14 \begin{small} ${\pm~1.15}$ \end{small} & 6.31 \begin{small} ${\pm~0.81}$ \end{small} & 8.96 \begin{small} ${\pm~2.15}$ \end{small}\\
  & $\text{Prompt-tuning}_{\textit{Manual}}$ & 85.60 \begin{small} ${\pm~0.81}$ \end{small} & 50.86 \begin{small} ${\pm~2.89}$ \end{small} & 52.15 \begin{small} ${\pm~0.98}$ \end{small} & 62.87 \begin{small} ${\pm~1.56}$ \end{small}\\
  & LM-BFF & 82.66 \begin{small} ${\pm~2.40}$ \end{small} & 56.03 \begin{small} ${\pm~0.65}$ \end{small} & 50.51 \begin{small} ${\pm~1.19}$ \end{small} & 63.07 \begin{small} ${\pm~1.41}$ \end{small}\\
  & RetroPrompt & 74.44 \begin{small} ${\pm~1.63}$ \end{small} & 36.49 \begin{small} ${\pm~1.07}$ \end{small} & 49.82 \begin{small} ${\pm~0.78}$ \end{small} & 53.58 \begin{small} ${\pm~1.16}$ \end{small}\\
  & KPT & 83.82 \begin{small} ${\pm~0.72}$ \end{small} & 61.83 \begin{small} ${\pm~0.83}$ \end{small} & 52.91 \begin{small} ${\pm~0.66}$ \end{small} & 66.19 \begin{small} ${\pm~0.74}$ \end{small}\\
  & $\text{S\textsc{ci}P\textsc{rompt}}$ & 84.71 \begin{small} ${\pm~0.89}$ \end{small} & 62.37 \begin{small} ${\pm~0.57}$ \end{small} & \textbf{53.65} \begin{small} ${\pm~0.22}$ \end{small} & 66.91 \begin{small} ${\pm~0.56}$ \end{small}\\
  & $\text{S\textsc{ci}P\textsc{rompt}}_{\textit{Soft}}$ & \textbf{85.96} \begin{small} ${\pm~0.60}$ \end{small} & \textbf{63.42} \begin{small} ${\pm~0.50}$ \end{small} & 52.41 \begin{small} ${\pm~0.30}$ \end{small} & \textbf{67.26} \begin{small} ${\pm~0.47}$ \end{small}\\
\midrule
\multirow{6}{*}{20} & $\text{Fine-tuning}_{\textit{SciBERT}}$ & 17.16 \begin{small} ${\pm~3.90}$ \end{small} & 3.53 \begin{small} ${\pm~0.86}$ \end{small} & 7.29 \begin{small} ${\pm~1.32}$ \end{small} & 9.33 \begin{small} ${\pm~2.03}$ \end{small}\\
  & $\text{Prompt-tuning}_{\textit{Manual}}$ & 87.76 \begin{small} ${\pm~0.70}$ \end{small} & 52.92 \begin{small} ${\pm~2.72}$ \end{small} & 54.32 \begin{small} ${\pm~0.89}$ \end{small} & 65.00 \begin{small} ${\pm~1.44}$ \end{small}\\
  & LM-BFF & 86.71 \begin{small} ${\pm~1.36}$ \end{small} & 60.90 \begin{small} ${\pm~0.22}$ \end{small} & 53.31 \begin{small} ${\pm~1.07}$ \end{small} & 66.97 \begin{small} ${\pm~0.88}$ \end{small}\\
  & RetroPrompt & 77.89 \begin{small} ${\pm~1.02}$ \end{small} & 41.79 \begin{small} ${\pm~0.81}$ \end{small} & 50.55 \begin{small} ${\pm~1.33}$ \end{small} & 56.74 \begin{small} ${\pm~1.05}$ \end{small}\\
  & KPT & 87.74 \begin{small} ${\pm~0.51}$ \end{small} & 66.25 \begin{small}${\pm~0.73}$ \end{small} & 54.67 \begin{small} ${\pm~0.43}$ \end{small} & 69.55 \begin{small} ${\pm~0.56}$ \end{small}\\
  & $\text{S\textsc{ci}P\textsc{rompt}}$ & \textbf{87.95} \begin{small} ${\pm~0.41}$ \end{small} & 66.59 \begin{small} ${\pm~0.64}$ \end{small} & \textbf{55.49} \begin{small} ${\pm~0.56}$ \end{small} & \textbf{70.01} \begin{small} ${\pm~0.54}$ \end{small}\\
  & $\text{S\textsc{ci}P\textsc{rompt}}_{\textit{Soft}}$ & 87.90 \begin{small} ${\pm~0.51}$ \end{small} & \textbf{66.86} \begin{small} ${\pm~0.46}$ \end{small} & 54.70 \begin{small} ${\pm~0.42}$ \end{small} & 69.82 \begin{small} ${\pm~0.46}$ \end{small}\\
\midrule
\multirow{6}{*}{50} & $\text{Fine-tuning}_{\textit{SciBERT}}$ & 27.50 \begin{small} ${\pm~9.48}$ \end{small} & 11.07 \begin{small} ${\pm~1.93}$ \end{small} & 12.02 \begin{small} ${\pm~2.22}$ \end{small} & 16.86 \begin{small} ${\pm~4.54}$ \end{small}\\
  & $\text{Prompt-tuning}_{\textit{Manual}}$ & 88.93 \begin{small} ${\pm~0.57}$ \end{small} & 60.63 \begin{small} ${\pm~1.32}$ \end{small} & 56.08 \begin{small} ${\pm~0.29}$ \end{small} & 68.55 \begin{small} ${\pm~0.73}$ \end{small}\\
  & LM-BFF & 87.94 \begin{small} ${\pm~0.56}$ \end{small} & 64.75 \begin{small} ${\pm~0.23}$ \end{small} & 54.97 \begin{small} ${\pm~0.69}$ \end{small} & 69.22 \begin{small} ${\pm~0.49}$ \end{small}\\
  & RetroPrompt & 83.14 \begin{small} ${\pm~0.63}$ \end{small} & 44.86 \begin{small} ${\pm~1.22}$ \end{small} & 53.04 \begin{small} ${\pm~0.73}$ \end{small} & 60.35 \begin{small} ${\pm~0.86}$ \end{small}\\
  & KPT & 88.93 \begin{small} ${\pm~0.37}$ \end{small} & 69.95 \begin{small} ${\pm~0.63}$ \end{small} & 56.50 \begin{small} ${\pm~0.81}$ \end{small} & 71.79 \begin{small} ${\pm~0.60}$ \end{small}\\
  & $\text{S\textsc{ci}P\textsc{rompt}}$ & \textbf{88.99} \begin{small} ${\pm~0.75}$ \end{small} & 69.89 \begin{small} ${\pm~0.63}$ \end{small} & \textbf{56.66} \begin{small} ${\pm~0.49}$ \end{small} & \textbf{71.85} \begin{small} ${\pm~0.62}$ \end{small}\\
  & $\text{S\textsc{ci}P\textsc{rompt}}_{\textit{Soft}}$ & 88.97 \begin{small} ${\pm~0.71}$ \end{small} & \textbf{70.15} \begin{small} ${\pm~0.52}$ \end{small} & 56.02 \begin{small} ${\pm~0.60}$ \end{small} & 71.71 \begin{small} ${\pm~0.61}$ \end{small}\\
\midrule[1pt] 
\multirow{1}{*}{Full Set} & $\text{Fine-tuning (Full) \textsuperscript{*}}$ & 90.71 & 54.58 & 53.74 & 66.34\\
\bottomrule[1.5pt]
\end{tabular}}
\caption{\label{few-shot}
Experimental results under few-shot settings. We report the mean accuracy (expressed in percentages \%) and standard deviation based on five iterations across five learning shots. Fine-tuning (Full)\textsuperscript{*} represents using a fully labeled training set. RetroPrompt experiments are only conducted in settings above five shots, as this method requires at least two labeled examples for model tuning.
}
\label{tab:fewshot}
\end{table*}

\subsection{Experimental Settings} \label{sec:4.2}
S\textsc{ci}P\textsc{rompt} is built upon the OpenPrompt framework \cite{ding2022openprompt}. We apply a consistent prompt template across all experiments (see Appendix~\ref{sec:appendixD} for more details). The experimental details are shown in Appendix~\ref{sec:appendixB}.


In the few-shot setting, we benchmark S\textsc{ci}P\textsc{rompt} alongside standard fine-tuning, simplified prompt-tuning (PT), and previous state-of-the-art text classification models, including LM-BFF \cite{gao2021making}, RetroPrompt \cite{chen2022decoupling}, and KPT \cite{hu2021knowledgeable}. Standard fine-tuning takes all labeled training examples as input to tuning an MLM for text classification. We take the final representation of the \verb|[CLS]| token as the output vector of the abstract \citep{cohan2020specter}. Standard PT with a manually defined verbalizer \cite{ding2022openprompt} only takes each lowercase topic name as a seed word for verbalization. We apply the same setting as in S\textsc{ci}P\textsc{rompt}, including a unified prompt template, MLM, and the model's hyper-parameters. KPT \cite{hu2021knowledgeable} applied external knowledge to enrich the verbalizer with additional word relevance and frequency filtering strategies. Our experiments use the same MLM (i.e., SciBERT) for equal comparison. Besides, training and validation examples per class \cite{ding2022openprompt, hu2021knowledgeable, wang2022automatic} are uniform during model tuning, conducting tests with 1, 5, 10, 20, and 50 shots across all datasets and reporting accuracy as an evaluation metric. We evaluate model performance across five random seeds to account for variability \cite{hu2021knowledgeable, ding2022openprompt}.

\begin{figure*}
  \centering
  \includegraphics[scale=0.135]{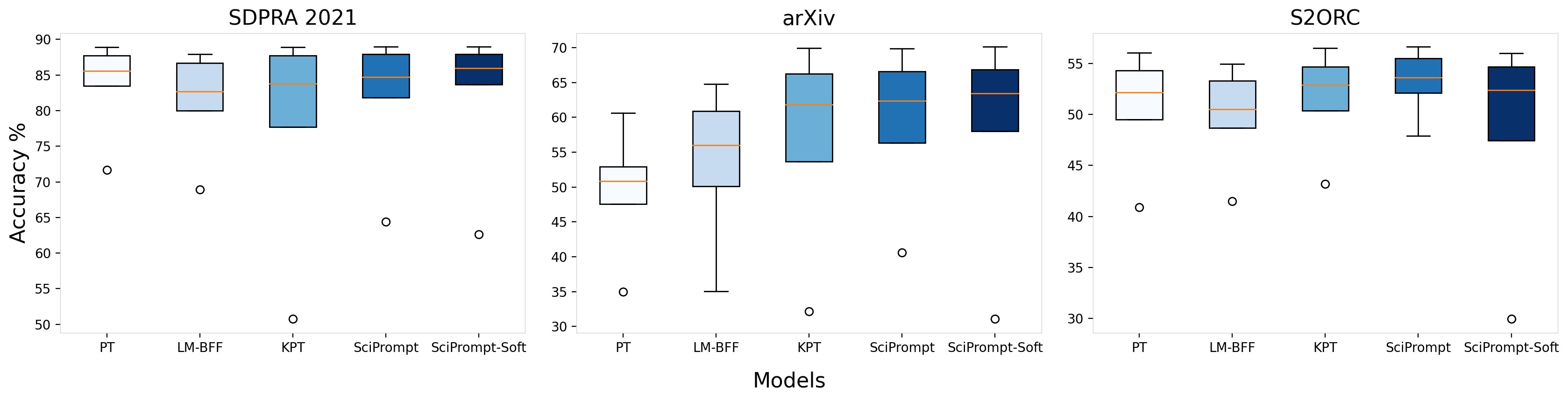}
  \caption{Performance comparison of few-shot methods over three datasets in Table~\ref{few-shot}. We report the mean accuracy of each setting. Our method shows high stability in the accuracy distribution compared to the considered baseline models.}
  \label{fig:line-chart}
\end{figure*}

For the zero-shot setting, we sample approximately 10\% of each dataset for testing, ensuring adequate representation for each topic. For broader model comparison, we introduce two additional models specific to the zero-shot scenario: SimPTC \cite{fei2022beyond} and NPPrompt \cite{zhao2023pre}. Moreover, we extend our evaluation to include Llama 2 \cite{touvron2023llama}, ChatGPT \cite{chatgpt}, and the latest Llama 3 \cite{llama3modelcard} using in-context learning for a broader range of comparisons. Random seeds are applied in KPT, which samples an unlabeled support set of 200 examples to calibrate label words. 

\section{Results and Analysis}

\subsection{Main Results} \label{sec:5.1}
We highlight the performance of S\textsc{ci}P\textsc{rompt} against baseline models across our three considered datasets in both few-shot and zero-shot settings, focusing on the fine-grained and cross-domain scientific text classification tasks. The experimental shown are listed in Table~\ref{tab:fewshot}. Results are averaged over five runs as the same as KPT \cite{hu2021knowledgeable} to counteract sampling randomness, reported as mean accuracy with standard deviation.

\textbf{Few-shot Results.} 
S\textsc{ci}P\textsc{rompt} achieves the best average accuracy on all three datasets for all settings. Specifically, S\textsc{ci}P\textsc{rompt} and S\textsc{ci}P\textsc{rompt}$_{\textit{Soft}}$ excel in low-data scenarios (e.g., one-shot and five-shot), particularly on arXiv and S2ORC, often outperforming baseline models. S\textsc{ci}P\textsc{rompt} also outperforms KPT by 8.93\% in the one-shot setting
and 2.83\% in the five-shot setting. As the number of training examples increases, the margin of improvement over baseline models narrows. Notably, S\textsc{ci}P\textsc{rompt} exceeds the full-set fine-tuning by an average of 0.57\%, 3.67\%, and 5.51\% with 10, 20, and 50 shots, respectively. Despite variability in performance improvements across different training sizes, our method consistently achieves the highest accuracy on arXiv and S2ORC across all configurations. Also, the standard deviation of all three datasets decreases as the number of input training examples increases across all three datasets.

Additionally, Figure~\ref{fig:line-chart} provides a comprehensive comparison of performances across all few-shot settings, ranging from one-shot to fifty-shot, for each dataset as outlined in Table~\ref{few-shot}. S\textsc{ci}P\textsc{rompt} consistently delivers high and stable accuracy across all three datasets compared to the baseline models. Particularly on S2ORC, S\textsc{ci}P\textsc{rompt} achieves a higher median accuracy and a narrower interquartile range, indicating more consistent performance across different few-shot scenarios. The S\textsc{ci}P\textsc{rompt}$_{\textit{Soft}}$ method shows high stability on the SDPRA 2021 dataset, while S\textsc{ci}P\textsc{rompt} is more effective in fine-grained datasets.

\begin{table}[h!]
    \footnotesize
    \scalebox{0.76}{
    \begin{tabular}{c|ccc|c}
    \midrule[1.5pt]
        \text{\textbf{Methods}} & \text{\textbf{SDPRA 2021}} & \text{\textbf{arXiv}} & \text{\textbf{S2ORC}} & \text{\textbf{Avg.}}\\
        \midrule[1pt]
        \text{Llama 2} & 62.04 & 26.98 & 40.30 & 43.11\\
        \text{Llama 3} & \textbf{81.15} & \textbf{54.87} & \textbf{49.58} & \textbf{61.87}\\
        \text{ChatGPT} & 79.43 & 54.51 & 46.95 & 60.30\\
        \midrule
        \text{PT} & 62.97 & 20.81 & 32.93 & \textbf{38.90} \\
        \text{SimPTC} & 15.79 & 3.25 & 11.35 & 10.13
        \\
        \text{NPPrormpt} & 35.00 & 13.98 & 37.23 & 28.74
        \\
        \text{LM-BFF} & \textbf{64.79} & 14.96 & 34.07 & 37.94\\
        \text{RetroPrompt} & 18.32 & 7.83 & 35.47 & 20.54 \\
        \text{KPT} & 41.50\begin{small}$\pm$3.00\end{small} & 20.83\begin{small}$\pm$0.18\end{small} & 38.42\begin{small}$\pm$0.66\end{small} & 33.58\begin{small}$\pm$1.28\end{small} \\
        $\text{S\textsc{ci}P\textsc{rompt}}$ & 51.97 & \textbf{22.28} & \textbf{41.30} & 38.52\\
        \midrule[1.5pt]
    \end{tabular}}
    \caption{Performance of zero-shot setting. Only KPT is reported through mean accuracy (\%) and standard deviation (\S\ref{sec:4.2}). We apply the same instruction for ChatGPT, Llama 2, and Llama 3 on the test sets.} 
    \label{tab4:zero-shot}
\end{table}
\textbf{Zero-shot Results.} Shown in Table~\ref{tab4:zero-shot}, the Llama 3 70B model leads in performance across all datasets. Nonetheless, S\textsc{ci}P\textsc{rompt} outperforms other baseline models, especially on arXiv and S2ORC, where it outperforms PT and KPT by margins of 1.47\% and 2.88\%, respectively. Meanwhile, LM-BFF leads among all baseline models on the SDPRA 2021 dataset. These results underscore the effectiveness of S\textsc{ci}P\textsc{rompt} in leveraging domain-specific knowledge for fine-grained scientific text classification, even in the absence of labeled training data. Llama 3's average accuracy exceeds S\textsc{ci}P\textsc{rompt} by 23.35\% and Llama 2's by 18.76\%. However, on the S2ORC dataset, S\textsc{ci}P\textsc{rompt} surpasses Llama 2. Note that S\textsc{ci}P\textsc{rompt}$_{\textit{Soft}}$ is not designed for zero-shot testing since it needs trainable tokens in the decoding layer during model tuning.


\begin{figure}[h!]
  \includegraphics[width=\linewidth]{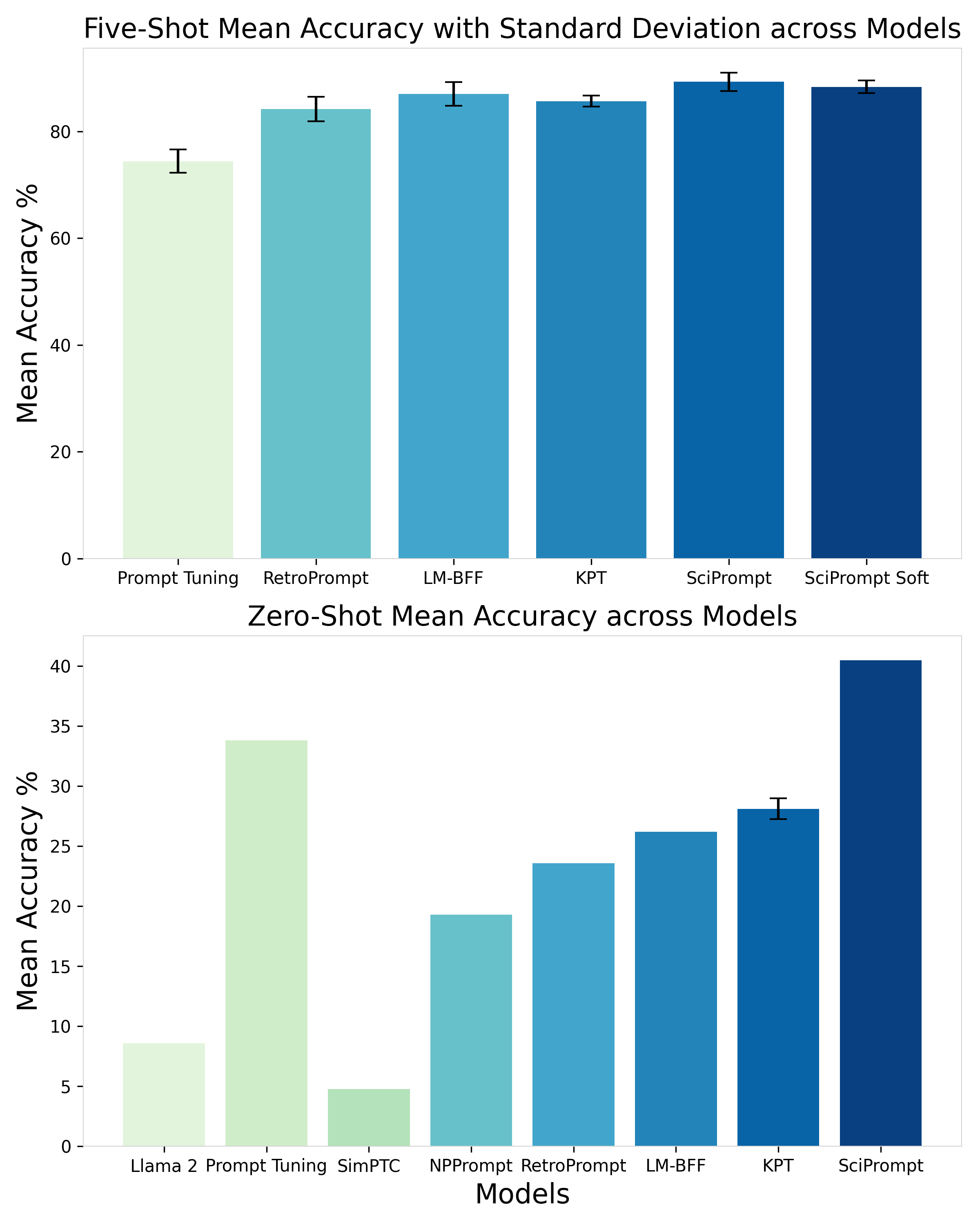}
  \caption{Model comparison through the Emerging NLP dataset under five-shot and zero-shot settings (\S\ref{emerging}).}
  \label{fig:emerging}
\end{figure}

\begin{table*}[h!]
    \centering
    \footnotesize
    \begin{tabular}{lccccc|c|c}
    \midrule[1.5pt]
        \textbf{Method} & \textbf{K=1} & \textbf{K=5} & \textbf{K=10} & \textbf{K=20} & \textbf{K=50} & \textbf{Avg.} & \textbf{Zero-shot}\\
        \midrule[1pt]
        $\text{KPT}$ & 32.18\begin{small}$\pm$1.08\end{small}&53.68\begin{small}$\pm$1.69\end{small}&61.83\begin{small}$\pm$0.83\end{small}&66.25\begin{small}$\pm$0.73\end{small}&69.95\begin{small}$\pm$0.63\end{small}&56.78&20.83\\
        \midrule
        S\textsc{ci}P\textsc{rompt} &40.57\begin{small}$\pm$1.60\end{small}&56.36\begin{small}$\pm$0.95\end{small}&62.37\begin{small}$\pm$0.57\end{small}&66.59\begin{small}$\pm$0.64\end{small}&69.89\begin{small}$\pm$0.63\end{small}&\textbf{59.16}&\textbf{22.28}\\
        $\hspace{2.4em}\text{w/o \textit{CL}}$&40.19\begin{small}$\pm$1.46\end{small}&55.84\begin{small}$\pm$0.98\end{small}&62.32\begin{small}$\pm$0.50\end{small}&66.45\begin{small}$\pm$0.61\end{small}&69.92\begin{small}$\pm$0.64\end{small}&58.94&21.87\\
        $\hspace{2.4em}\text{w/o \textit{SS}}$&38.70\begin{small}$\pm$0.86\end{small}&55.19\begin{small}$\pm$0.80\end{small}&62.48\begin{small}$\pm$0.59\end{small}&66.70\begin{small}$\pm$0.77\end{small}&69.73\begin{small}$\pm$1.01\end{small}&58.56&6.17\\
        \hspace{0.7em}w/o \textit{SS}+\textit{CL}&38.36\begin{small}$\pm$0.86\end{small}&54.76\begin{small}$\pm$0.86\end{small}&62.25\begin{small}$\pm$0.56\end{small}&66.54\begin{small}$\pm$0.81\end{small}&69.86\begin{small}$\pm$0.92\end{small}&58.35&5.62\\
        \hspace{0.5em}w/o \textit{FL}+\textit{CL}&29.77\begin{small}$\pm$0.74\end{small}&50.13\begin{small}$\pm$0.88\end{small}&59.57\begin{small}$\pm$0.97\end{small}&65.77\begin{small}$\pm$0.47\end{small}&69.55\begin{small}$\pm$0.70\end{small}&54.96&3.77\\
        \midrule
        S\textsc{ci}P\textsc{rompt}$_{\textit{Soft}}$&31.06\begin{small}$\pm$1.74\end{small}&58.01\begin{small}$\pm$0.94\end{small}&63.42\begin{small}$\pm$0.50\end{small}&66.86\begin{small}$\pm$0.46\end{small}&70.15\begin{small}$\pm$0.52\end{small}&57.90& - \\
        \hspace{2.4em}w/o \textit{CL}&38.65\begin{small}$\pm$0.90\end{small}&58.33\begin{small}$\pm$1.62\end{small}&63.64\begin{small}$\pm$0.66\end{small}&67.05\begin{small}$\pm$0.55\end{small}&70.41\begin{small}$\pm$0.56\end{small}&59.62& - \\
        \hspace{2.4em}w/o \textit{SS}&41.49\begin{small}$\pm$1.38\end{small}&58.36\begin{small}$\pm$0.99\end{small}&63.70\begin{small}$\pm$0.75\end{small}&67.26\begin{small}$\pm$0.75\end{small}&70.20\begin{small}$\pm$0.21\end{small}&\textbf{60.20}& - \\
        \hspace{0.7em}w/o \textit{SS}+\textit{CL}&42.22\begin{small}$\pm$1.32\end{small}&57.72\begin{small}$\pm$1.46\end{small}&63.53\begin{small}$\pm$0.57\end{small}&67.03\begin{small}$\pm$0.78\end{small}&70.35\begin{small}$\pm$0.49\end{small}&60.17& - \\
        \hspace{0.5em}w/o \textit{FL}+\textit{CL} &37.50\begin{small}$\pm$1.31\end{small}&57.66\begin{small}$\pm$1.49\end{small}&63.63\begin{small}$\pm$0.49\end{small}&67.13\begin{small}$\pm$0.93\end{small}&70.24\begin{small}$\pm$0.49\end{small}&59.23& -\\
    \bottomrule[1.5pt]
    \end{tabular}
    \caption{
Ablation study of S\textsc{ci}P\textsc{rompt} for mean accuracy and standard deviation for the arXiv dataset under few-shot and zero-shot settings.
}
\label{tab5:ablation}
\end{table*}

\subsection{Emerging Topics Classification} \label{emerging}
To assess our method's effectiveness in classifying emerging scientific topics, we manually collect a dataset centered around recent developments in the field of NLP, drawing inspiration from \citet{ahmad2024forc4cl}. Specifically, we first extract NLP topics from Taxonomy4CL\footnote{\url{https://github.com/DFKI-NLP/Taxonomy4CL}}, focusing on topics that have emerged since 2000, as identified through Semantic Scholar\footnote{\url{https://www.semanticscholar.org/}}. We then select scientific articles published after 2019 that are beyond the knowledge cutoff of the SciBERT model. For each selected topic, we gather 30 abstracts, applying the same random seeds for few-shot experiments as those introduced in Table~\ref{few-shot}. We create a new dataset named \textbf{Emerging NLP} by collecting 21 fine-grained NLP-related topics and their corresponding abstracts. Appendix~\ref{sec:appendixC} provides detailed dataset statistics and topic examples. Figure~\ref{fig:emerging} compares the performance of various baseline models. Notably, S\textsc{ci}P\textsc{rompt} exceeds the performance of the Llama 2 70B model by 31.91\% and outperforms the PT method by 6.67\% in the zero-shot setting. Overall, our method outperforms all state-of-the-art methods in classifying emerging scientific topics, especially in the zero-shot setting, highlighting our method's efficacy in highly low-resource scenarios.

\subsection{Ablation Study} \label{sec:5.3}
Our ablation study on the arXiv dataset (Table~\ref{tab5:ablation}) demonstrates the advantages of our models over KPT, with a 1.45\% increase in zero-shot accuracy. S\textsc{ci}P\textsc{rompt} and S\textsc{ci}P\textsc{rompt}$_{\textit{Soft}}$ outperform KPT by 2.38\% and 3.42\%, respectively, in terms of average accuracy under the few-shot setting. We examine the impact of removing full-size calibration (``w/o \textit{CL}''), semantic scores (``w/o \textit{SS}''), and both (``w/o \textit{SS}+\textit{CL}''), finding that both components improve the performance, especially in the zero-shot setting where their absence lowers accuracy by 0.41\% (``w/o \textit{CL}'') and 16.11\% (``w/o \textit{SS}'') compared to S\textsc{ci}P\textsc{rompt}, underlining the critical role of \textit{SS} in bolstering the model's effectiveness.

Interestingly, S\textsc{ci}P\textsc{rompt}$_{\textit{Soft}}$ performs better without \textit{SS} than when both components are included. Removing both \textit{SS} and \textit{CL} yields the best 1-shot performance, suggesting that less intervention optimizes model tuning in low-data contexts. Furthermore, comparing setups without pre-filtering and calibration (``w/o \textit{FL}+\textit{CL}'') to those with pre-filtering shows an accuracy increase by 3.39\% and 0.94\% for S\textsc{ci}P\textsc{rompt} and S\textsc{ci}P\textsc{rompt}$_{\textit{Soft}}$ respectively, highlighting the effectiveness of pre-filtering of augmented verbalizer for text classification. The ablation studies of SDPRA and S2ORC shows the same pattern as on arXiv.

\subsection{Model Tuning Efficiency} \label{sec:5.5} 
Table~\ref{tab:efficiency} shows that S\textsc{ci}P\textsc{rompt}$_{\textit{Soft}}$ reduces GPU memory usage by 16.5 percentage points (p.p.) for SDPRA 2021, 38 p.p. for arXiv, and 46.2 p.p. for S2ORC compared to S\textsc{ci}P\textsc{rompt}'s full-size label term calibration. Although S\textsc{ci}P\textsc{rompt} achieves higher average accuracy rates in the few-shot setting on the S2ORC dataset (see Table~\ref{tab7:cali} in Appendix~\ref{sec:appendixA}), S\textsc{ci}P\textsc{rompt}$_{\textit{Soft}}$ outperforms S\textsc{ci}P\textsc{rompt} on SDPRA 2021 and arXiv, suggesting that S\textsc{ci}P\textsc{rompt}$_{\textit{Soft}}$ can achieve competitive results with less GPU usage. Moreover, while ChatGPT and Llama 2 exhibit superior performance in the zero-shot setting, as shown in Table~\ref{tab4:zero-shot}, it is worth noting that these language models are either mainly for commercial use or require substantial GPU resources, incurring higher costs or more time. For instance, for the S2ORC dataset, our method not only cuts down the combined training and testing (inference) time by 93 p.p. compared to Llama 2 70B but also enhances accuracy by 1 p.p. over Llama 2, highlighting the efficiency and effectiveness of our approach.

\begin{table}
\centering
\small
\begin{tabular}{cccc}
\toprule[1.5pt]
\textbf{Method} & \textbf{SDPRA 2021} & \textbf{arXiv} & \textbf{S2ORC}\\
\midrule[1pt]
S\textsc{ci}P\textsc{rompt}\\
\midrule
w/o CL & 9.5\% & 12.5\% & 12.4\% \\
w/ CL & 29.3\% & 51.2\% & 59.6\% \\
\midrule
S\textsc{ci}P\textsc{rompt}$_{\textit{Soft}}$\\
\midrule
w/o CL & 12.3\% & 12.3\% & 12.3\% \\
w/ CL & 12.8\% & 13.2\% & 13.4\% \\
\bottomrule[1.5pt]
\end{tabular}
\caption{
The usage percentage of GPU memory during model tuning.
}
\label{tab:efficiency}
\end{table}

\section{Conclusion}
We introduced a knowledge-enhanced, prompt-based fine-tuning framework for fine-grained scientific text classification using minimally or no labeled abstracts. Acknowledging the complexity of domain knowledge within scientific literature, we employed a prompt-tuned MLM augmented with domain knowledge injection and semantic filtering. This approach enables the automatic extraction of domain-specific phrases and their integration into a weighted verbalizer for topic projection. Our findings highlight the effectiveness of our methods over existing state-of-the-art models and standard full-set fine-tuning, particularly for emerging topic classification and scenarios requiring high levels of topic granularity. Notably, S\textsc{ci}P\textsc{rompt} demonstrates competitive accuracy compared to the advanced Llama 2 70B model in the zero-shot setting, showing its potential to categorize scholarly topics with a lightweight and efficient approach. 

\section{Limitations}
Our study's limitations are as follows: 
1) Our external knowledge sources are limited to two non-scientific domain databases for retrieving topic words, potentially missing fine-grained scientific terminologies. Despite the challenge of identifying a universally applicable, cross-domain, scientific knowledge resource, future efforts should aim to discover more precise terminology databases \cite{han2020wikicssh}.
2) We focus solely on a multi-class classification task and exclude abstracts that span multiple scientific sub-domains. Advancing towards a multi-label classification system capable of identifying publications across various domains would enhance the robustness of our approach.
3) Although S\textsc{ci}P\textsc{rompt} and S\textsc{ci}P\textsc{rompt}$_{\textit{Soft}}$ surpassed baseline methods during evaluation, the enhancements are modest, and results fluctuate, particularly with an increase in labeled training data. Further investigation into the causes of these minimal gains as well as more comprehensive, interpretable experiments are needed to better understand and improve the model performance. 
4) We only used classification accuracy and standard deviation as model evaluation metrics. The experimental results can change when using other metrics (e.g., Micro F1 and Macro F1). Additionally, while the standard deviation of our methods shrinks as the number of training examples increases, one could do statistical significance testing to draw robust conclusions by comparing system performance against baseline models.

\section{Ethics Statement}
The datasets and MLM employed in our study are publicly accessible and extensively utilized in the research community. To enhance the quality of our data, we applied heuristic filtering to exclude short-length abstracts across these datasets, acknowledging that this process may impact experimental accuracy. Our methodology includes extracting data from external knowledge bases via public APIs. Furthermore, as we used MLMs as the foundation of our approach, it is essential to note that the predictive behavior of these models can be challenging to regulate due to the implicit knowledge embedded within the MLMs, which is difficult to decode explicitly. Therefore, caution should be exercised when adapting our method to other tasks, especially in the context of text classification through prompting.

\clearpage
\appendix

\begin{table}[h!]
\footnotesize
\scalebox{0.85}{
\begin{tabular}{ccccc}
\toprule[1.5pt]
\textbf{Datasets} & \textbf{\#Abstracts} & \textbf{\makecell{\# Classes }} & \textbf{Avg. Length} & \textbf{Test}\\
\midrule[1pt]
arXiv & 55300 & 53 (sub) & 129 & 5300\\
\midrule
SDPRA 2021 & 28000 & 7 & 155 & 2800\\
\midrule
S2ORC & 65700 & 19 & 136 & 5700\\
\midrule
Emerging NLP & 630 & 21 & 227 & 420\\
\bottomrule[1.5pt]
\end{tabular}}
\caption{
Datasets Statistics. \#Abstracts represents the total number of labeled abstracts, including train and test sets. Emerging NLP dataset is for five-shot and zero-shot settings only.
}
\label{tab:datasets-stat}
\end{table}

\section{Experimental Details}
\label{sec:appendixB}
All models use the maximum input length of 256 tokens over 5 epochs, using the same hyper-parameters as KPT \cite{hu2021knowledgeable}, with a learning rate of 3e-5 and a batch size of 5. The experiments are performed on a 32 GB Tesla V100 GPU.

In few-shot setting, we apply the same backbone MLM for all experiments, with the exception of RetroPrompt \cite{chen2022decoupling}. RetroPrompt only supports RoBERTa-based models and requires at least two examples per class for model tuning. Therefore, we apply \texttt{
roberta-base} as base model for RetroPrompt and only conduct experiments with more than five shots. 

The main distinction between S\textsc{ci}P\textsc{rompt} and S\textsc{ci}P\textsc{rompt}$_{\textit{Soft}}$ lies in the verbalization, as discussed in Section~\ref{sec:3.6}. Unlike S\textsc{ci}P\textsc{rompt}, which uses single label term projection, S\textsc{ci}P\textsc{rompt}$_{\textit{Soft}}$ employs a vector-based mapping method to represent each filtered set of label terms. 

In zero-shot setting, we include ChatGPT\footnote{\url{https://openai.com/chatgpt}}, open-sourced Llama 2\footnote{\url{https://llama.meta.com/}}, and the latest Llama 3\footnote{\url{https://ai.meta.com/blog/meta-llama-3/}} for zero-shot classification using the same instruction. For ChatGPT, we use \verb|gpt-3.5-turbo-instruct|, which contains 175 million model parameters developed by OpenAI. We apply \verb|llama-2-70b-chat| and \verb|meta-llama-3-70b-instruct| as the backbone models for Llama 2 and Llama 3 respectively through the Replicate API\footnote{\url{https://replicate.com/}}. We additionally investigate the classification performance of the Llama 2 models with 7B and 13B parameters under the zero-shot setting. However, their outputs are not coherent with the predefined class label sets and often include redundant information, making the calculation of prediction accuracy unreliable. Therefore, we only conduct experiments of the Llama family on the 70B models.

\begin{table*}[!ht]
    \centering
    \footnotesize
    \begin{tabular}{lccccc|c|c}
    \midrule[1.5pt]
        \textbf{Paradim} & \textbf{K=1} & \textbf{K=5} & \textbf{K=10} & \textbf{K=20} & \textbf{K=50} & \textbf{Avg.} & \textbf{Zero-Shot}\\
        \midrule[1pt]
        S\textsc{ci}P\textsc{rompt} (SDPRA)\\
        \midrule        
        $\hspace{2.2em}\text{w/o \textit{FL}}$&45.23&73.20&81.61&87.40&88.94&75.28&25.56\\
        $\hspace{2.2em}\text{w/ \textit{FL}}$&61.25&81.33&84.67&87.78&89.05&80.83&34.98\\
        \hspace{2.2em}w/ \textit{CL}&63.56&81.57&84.62&88.02&89.02&\textbf{81.36}&\textbf{51.40}\\
        \midrule        
        S\textsc{ci}P\textsc{rompt}$_{\textit{Soft}}$ (SDPRA)\\
        \midrule
        \hspace{2.2em}w/o \textit{FL}&55.00&80.62&83.84&87.91&88.69&79.21& - \\
        \hspace{2.2em}w/ \textit{FL}&65.53&83.43&85.26&88.13&88.97&\textbf{82.26}& - \\
        \hspace{2.2em}w/ \textit{CL}&64.92&81.78&85.46&87.79&89.14&81.82& - \\
    \midrule[1.5pt]
        S\textsc{ci}P\textsc{rompt} (arXiv)\\
        \midrule        
        $\hspace{2.2em}\text{w/o \textit{FL}}$&29.77&50.13&59.57&65.77&69.55&54.96&3.77\\
        $\hspace{2.2em}\text{w/ \textit{FL}}$&38.36&54.76&62.25&66.54&69.86&58.35&5.62\\
        \hspace{2.2em}w/ \textit{CL}&38.70&55.19&62.48&66.70&69.73&\textbf{58.56}&\textbf{6.17}\\
        \midrule        
        S\textsc{ci}P\textsc{rompt}$_{\textit{Soft}}$ (arXiv)\\
        \midrule
        \hspace{2.2em}w/o \textit{FL}&37.50&57.66&63.63&67.13&70.24&59.23& - \\
        \hspace{2.2em}w/ \textit{FL}&42.22&57.72&63.53&67.03&70.35&60.17& - \\
        \hspace{2.2em}w/ \textit{CL}&41.49&58.36&63.70&67.26&70.20&\textbf{60.20}& - \\
    \midrule[1.5pt]
        S\textsc{ci}P\textsc{rompt} (S2ORC)\\
        \midrule        
        $\hspace{2.2em}\text{w/o \textit{FL}}$&41.27&49.22&52.69&55.30&56.31&50.96&25.25\\
        $\hspace{2.2em}\text{w/ \textit{FL}}$&46.00&51.23&53.43&55.25&56.15&52.41&26.11\\
        \hspace{2.2em}w/ \textit{CL}&47.55&51.85&53.52&55.32&56.67&\textbf{52.98}&\textbf{40.79}\\
        \midrule        
        S\textsc{ci}P\textsc{rompt}$_{\textit{Soft}}$ (S2ORC)\\
        \midrule
        \hspace{2.2em}w/o \textit{FL}&42.35&50.10&51.89&54.52&56.17&51.01& - \\
        \hspace{2.2em}w/ \textit{FL}&46.33&50.24&52.83&54.76&56.17&52.07& - \\
        \hspace{2.2em}w/ \textit{CL}&46.34&51.09&53.02&54.59&55.82&\textbf{52.17}& - \\
    \midrule[1.5pt]
    \end{tabular}
    \caption{
Performance comparison under various number of label terms in the verbalizer. We report the mean accuracy after five runs for each shot.
}
\label{tab7:cali}
\end{table*}

\section{Datasets and Examples of Domain Topic Categories}
\label{sec:appendixC}
We present a more detailed introduction to datasets used for our experiments.

\textbf{SDPRA 2021} contains topics of scientific articles from the field of computer science, consisting of abstracts sourced from arXiv and categorized under one of seven predefined domain labels. We combined the training and validation sets, reallocating them into new training (90\%) and validation (10\%) sets.

\textbf{arXiv} includes abstracts sourced from the arXiv website collected by \citet{meng2019weakly}, categorized into 53 sub-categories and 3 parent categories (i.e., Math, Physics, and CS). We select 100 samples for each category as test set.

\textbf{S2ORC} includes academic papers across 19 disciplines. We filter abstracts to those with a single discipline label from the 2023-11-07 release through the Semantic Scholar Public API\footnote{\url{https://www.semanticscholar.org/product/api}}.

\textbf{Emerging Topics of NLP} encompasses 21 newly developed research fields within the broader category of Computation and Language\footnote{\url{https://arxiv.org/list/cs.CL/recent}}. We collect 30 examples for each topic, assigning five instances for training and another five for validation. The rest of the examples are used for testing.

In our experiments, abstracts shorter than 30 tokens were excluded to remove invalid abstracts, leading to final training and test sizes of 25,110 and 2,790 for SDPRA, 49,300 and 5,300 for arXiv, 60,000 and 5,700 for S2ORC, and 210 and 420 for Emerging NLP. We used sub-categories for arXiv and parent categories for both SDPRA and S2ORC in text classification tasks. Detailed class labels for each dataset are presented in Table~\ref{tab8:data-categories}. We report parent and sub-categories of four datasets.

\section{Experiments of Various Verbalizer Sizes}
\label{sec:appendixA}
\begin{figure}[h!]
  \includegraphics[width=\linewidth]{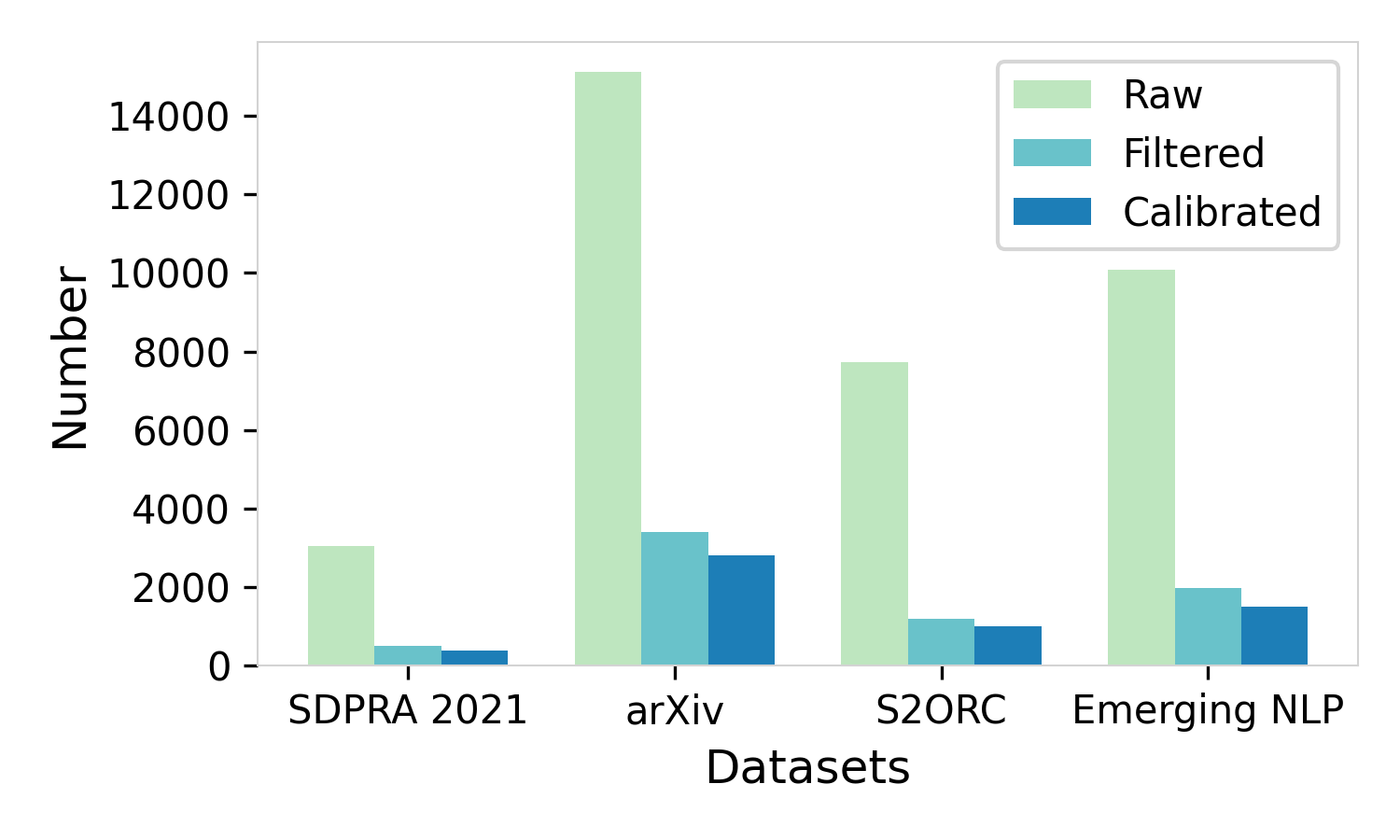}
  \caption{Various numbers of label terms across four datasets under three phrases.}
  \label{fig:bar-chart}
\end{figure}

As presented in Table~\ref{tab7:cali}, we document the performance metrics across various verbalizer sizes following the configurations outlined in Figure~\ref{fig:bar-chart}. We report the mean accuracy for each setting. The findings indicate that the model's performance is enhanced across all scientific domain text classification datasets in both few-shot and zero-shot scenarios, attributable to implementing more sophisticated label term filtering techniques.

\section{Calibration of Domain Knowledge} \label{calibration}
Figure~\ref{fig:bar-chart} compares the verbalizer label term counts across datasets: ``Raw'' reflects the initial count after knowledge retrieval from two KBs (\S\ref{sec:3.2}); ``Filtered'' shows counts post-semantic filtering (\S\ref{sec:3.4}), reducing terms by 84\%, 77\%, and 85\%; ``Calibrated'' involves removing low-likelihood terms before model tuning. Appendix~\ref{sec:appendixA} and Table~\ref{tab7:cali} reveal that NLI filtering and calibration enhance the model's accuracy in few-shot and zero-shot settings, linking domain-relevant phrases in the verbalizer to improve the classification performance.

\begin{table*}
    \footnotesize
    \centering
    \begin{tabular}{cccccc}
    \midrule[1.5pt]
        \text{\textbf{Datasets}} & \text{\textbf{$\mathcal{M}_{ce}$ < 0.1}} & \text{\textbf{$\mathcal{M}_{ce}$ < 0.3}} & \text{\textbf{$\mathcal{M}_{ce}$ < 0.6}} & \text{\textbf{$\mathcal{M}_{ce}$ < 0.9}} & \text{\textbf{$\mathcal{M}_{ce}$ > 0.9}}\\
        \midrule[1pt]
        \text{SDPRA} & 495 & 501 & 514 & 531 & 738\\
        \text{arXiv} & 3,384 & 3,477 & 3,553 & 3,678 & 5,646\\
        \text{S2ORC} & 1,182 & 1,216 & 1,239 & 1,283 & 1,771\\
        \midrule[1.5pt]
    \end{tabular}
    \caption{The number of filtered label terms applying various thresholds.} 
    \label{label-terms-threshold}
\end{table*}

\begin{table*}[h!]
    \footnotesize
    \centering
    \begin{tabular}{ccccccc}
    \midrule[1.5pt]
        \text{\textbf{Cross-Encoder}} & \text{\textbf{$\mathcal{M}_{be}$ < 0.5}} & \text{\textbf{$\mathcal{M}_{be}$ > 0.5}} & \text{\textbf{$\mathcal{M}_{be}$ > 0.6}} & \text{\textbf{$\mathcal{M}_{be}$ > 0.7}} & \text{\textbf{$\mathcal{M}_{be}$ > 0.8}} & 
        \text{\textbf{$\mathcal{M}_{be}$ > 0.9}}\\
        \midrule[1pt]
        \text{$\mathcal{M}_{ce}$ < 0.1} & 64.18\begin{small}$\pm$5.83\end{small} & 64.42\begin{small}$\pm$3.64\end{small} & 65.94\begin{small}$\pm$4.84\end{small} & 
        64.69\begin{small}$\pm$5.24\end{small} & 
        64.79\begin{small}$\pm$4.19\end{small} & 
        66.67\begin{small}$\pm$3.90\end{small}\\
        \midrule[1.5pt]
    \end{tabular}
    \caption{Ablation study of S\textsc{ci}P\textsc{rompt} in various $\mathcal{M}_{be}$ values under the fixed $\mathcal{M}_{ce}$ using the SDPRA 2021 dataset.} 
    \label{cross-encoder-threshold}
\end{table*}

\begin{figure}[h!]
  \includegraphics[scale=0.39]{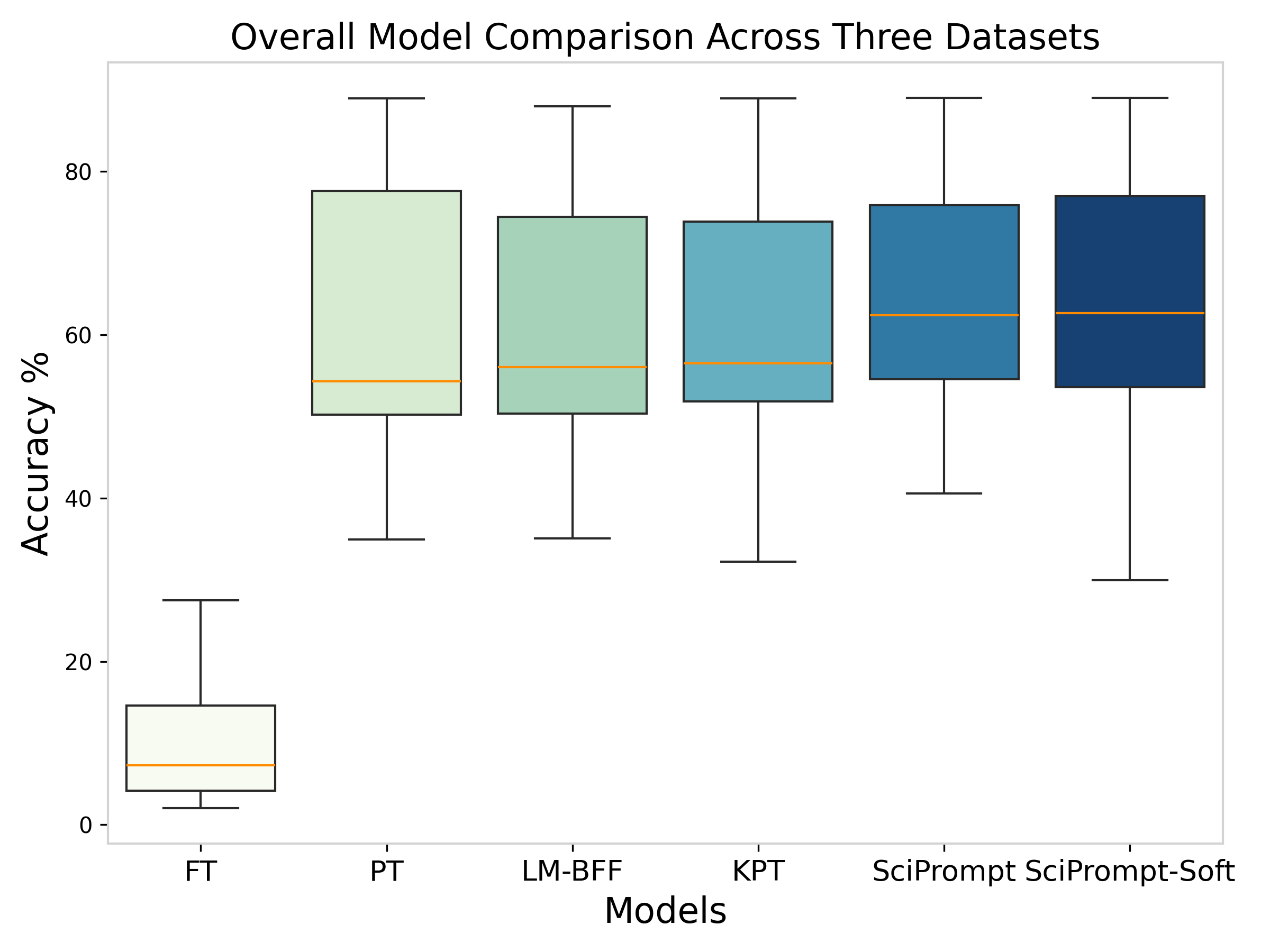}
  \caption{Box chart for all methods in the few-shot setting over three datasets.}
  \label{fig:box-chart}
\end{figure}

\section{Overall Model Performance Analysis}
We present an overview comparison of the results from Table~\ref{tab:fewshot} across all three datasets (i.e., SDPRA 2021, arXiv, and S2ORC) in Figure~~\ref{fig:box-chart}. Overall, S\textsc{ci}P\textsc{rompt} exhibits the most stable performance compared to other baseline methods. Notably, S\textsc{ci}P\textsc{rompt} consistently outperforms the state-of-the-art model KPT across all three datasets. In contrast, S\textsc{ci}P\textsc{rompt}$_{\textit{Soft}}$ demonstrates variability and inconsistency compared with S\textsc{ci}P\textsc{rompt} while showing a similar median accuracy. We exclude the RetroPrompt method from this comparison due to its inability to perform in the one-shot setting.

\section{Knowledge-Retrieval Threshold Selection}
\label{threshold}
As we introduced in Section~\ref{sec:3.4}, during the label term filtering stage, we employ a bi-encoder for $\mathcal{M}_{be}$ and a cross-encoder for $\mathcal{M}_{ce}$ calculation. In our experimentation, a higher $\mathcal{M}_{be}$ score indicates a more notable similarity between the topic labels and the retrieved label terms, thus enhancing the relevance of the selected terms. Conversely, a lower $\mathcal{M}_{ce}$ score signifies higher relevance during the re-ranking stage. Our analysis of the SDPRA dataset reveals that $\mathcal{M}_{ce}$ scores predominantly clustered above 0.9 and below 0.1. Consequently, the median value of $\mathcal{M}_{ce}$ exerts minimal influence on the final Verbalization process. Even when reducing the threshold of $\mathcal{M}_{ce}$ to 0.5, only a marginal difference in the number of selected label terms across various $\mathcal{M}_{ce}$ scores within the range of 0.1 to 0.9 is observed (Table~\ref{label-terms-threshold}).

We also explored the impact of different $\mathcal{M}_{be}$ values under the fixed $\mathcal{M}_{ce}$ score (0.1) to assess performance variations of S\textsc{ci}P\textsc{rompt} in the 1-shot setting through the SDPRA dataset. Our findings indicate that while $\mathcal{M}_{be}$ > 0.9 yields the optimal performance, $\mathcal{M}_{be}$ > 0.5 kept the lowest standard deviation (Table~\ref{cross-encoder-threshold}). Consequently, we assume setting $\mathcal{M}_{be}$ > 0.5 as the filtering threshold is more stable across different experimental conditions.

\begin{table}[h!]
    \footnotesize
    \begin{tabular}{cccc}
    \midrule[1.5pt]
        \text{\textbf{Bi-Encoder}} & \text{\textbf{$\mathcal{M}_{ce}$ < 0.1}} & \text{\textbf{$\mathcal{M}_{ce}$ < 0.5}} & \text{\textbf{$\mathcal{M}_{ce}$ > 0.5}}\\
        \midrule[1pt]
        \text{$\mathcal{M}_{be}$ > 0.5} & 64.42\begin{small}$\pm$3.64\end{small} & 63.80\begin{small}$\pm$5.11\end{small} & 34.86\begin{small}$\pm$6.61\end{small}\\
        \midrule[1.5pt]
    \end{tabular}
    \caption{Ablation study of S\textsc{ci}P\textsc{rompt} in various $\mathcal{M}_{ce}$ values under the fixed $\mathcal{M}_{be}$ using the SDPRA 2021 dataset.} 
    \label{biencoder-threshold}
\end{table}

To further validate our choices, we conducted experiments of S\textsc{ci}P\textsc{rompt} with varying $\mathcal{M}_{ce}$ values under the 1-shot setting using the SDPRA dataset while maintaining a constant $\mathcal{M}_{be}$ threshold of 0.5. Notably, performance consistently improved and the standard deviation is stable when $\mathcal{M}_{ce}$ is set below 0.1 (Table~\ref{biencoder-threshold}). Therefore, we adopted $\mathcal{M}_{ce}$ = 0.1 as the filtering threshold.

\section{Prompt Templates of LLMs}
\label{sec:appendixD}

\begin{tcolorbox}[enhanced,boxrule=1pt,colback=white,colframe=black,boxsep=5pt,arc=4pt]
\noindent \textbf{Cloze-Based Prompt Template of MLM}\\

\noindent \verb|Abstract|. The field of this study is related to: \verb|[MASK]|.
\end{tcolorbox}

Above is the cloze-based prompt template we applied for all MLM prompt-based fine-tuning tasks. We also explored various prompt templates as introduced by \cite{hu2021knowledgeable, gao-etal-2021-making, you-etal-2024-uiuc} to evaluate performance variations using the SDPRA 2021 dataset, where the results are found to be similar. Note that our method focuses on improving domain-related verbalization process rather than creating diverse prompts for model tuning.

As detailed in Section~\ref{sec:5.1}, we used ChatGPT, Llama 2, and Llama 3 to perform the task of scientific text classification guided by specific instructions. The same instructions were applied to all LLMs to infer the topics from scientific abstracts. We employed a distinct task-oriented \cite{you-etal-2024-beyond} prompt from that used with MLMs due to our observation that the original prompt from S\textsc{ci}P\textsc{rompt} fails to yield relevant field names, given the LLMs' limitations in comprehension. Consequently, we crafted a more elaborate set of instructions to direct the LLMs in classifying topics, employing a projection of pre-defined class names similar to those used in the verbalization.

\begin{tcolorbox}[enhanced,boxrule=1pt,colback=white,colframe=black,boxsep=5pt,arc=4pt]
\noindent \textbf{Instructions of LLMs}\\

\noindent Based on the given article's abstract, please classify the abstract to a specific field of study. Only select field words from the following field words I provided. Only select one field name as output for each abstract. Your output should be all in lower cases.\texttt{\textbackslash n}\\

\noindent \textbf{Field Words List:} logic in computer science, distributed computing, software engineering, data structures and algorithms, computational linguistics... \texttt{\textbackslash n}\\

\noindent \textbf{Abstract:} For the purpose of developing applications for Post-K at an early stage, RIKEN has developed a post-K processor simulator. This simulator is based on the general-purpose processor simulator gem5...\texttt{\textbackslash n}\\

\noindent \textbf{Field of Study}: 
\end{tcolorbox}






The ``Field Words List'' represents the original class names in the dataset. We concatenate the above instructions to LLMs and extract the predictions that appear after ``Field of Study:'' to evaluate the classification performance. 




\section{Examples of Retrieved Label Terms}
In Table~\ref{retrieval-examples}, we report some cases of filtered label terms using the KBs we introduced in Section~\ref{sec:3.2} through four datasets we apply for this work. 

\begin{table*}
\centering
\scalebox{0.98}{
\begin{tabular}{c|c|c}
\toprule[1.5pt]
\textbf{Datasets} & \textbf{Class Labels} & \textbf{Filtered Label Terms}\\
\midrule[1pt]
\multirow{7}{*}{arXiv} & Databases & \makecell{document-oriented database, \\hierarchical database, \\database management system, \\ object database, \\database application}\\
\cline{2-3} 
& Accelerator Physics & \makecell{accelerator physics, \\particle accelerator, particle beam, \\velocity,accelerator} \\
\cline{2-3} 
& Group Theory & \makecell{symmetry group, group homomorphism, \\representation theory of finite groups, \\compact lie group} \\
\midrule
\multirow{1}{*}{SDPRA 2021} & Cryptography & \makecell{cryptographers, secure communication, \\ciphertext, cryptanalytics, \\cryptographers, secure communication, \\data encryption standard}\\
\midrule
\multirow{5}{*}{S2ORC} & Political Science & \makecell{political behavior, aspects, \\politics, elections, \\practical politics, \\american political science, \\constitutions, governing}\\
\cline{2-3} 
& Psychology & \makecell{psychological science, \\mental condition, mental state, \\mental function, psychological state, \\psychological condition} \\
\midrule
\multirow{5}{*}{Emerging NLP} & Large Language Models (LLMs) & \makecell{bert, semi-supervised learning, \\chain-of-thought prompting, \\encoding, lstm}\\
\cline{2-3} 
& Recurrent Neural Networks (RNNs) & \makecell{tensor, language modeling, \\generative model, \\feedforward neural networks, \\gated recurrent unit} \\
\bottomrule[1.5pt]
\end{tabular}}
\caption{
Examples of filtered label terms in four datasets (\S\ref{sec:3.4}). 
}
\label{retrieval-examples}
\end{table*}

\begin{table*}[!ht]
\centering
\footnotesize
\scalebox{0.95}{
\begin{tabular}{c|c|c}
\toprule[1.5pt]
\textbf{Datasets} & \textbf{Parent-category} & \textbf{Sub-category}\\
\midrule[1pt]
\multirow{17}{*}{arXiv} & Math (25) & \makecell{numerical analysis, algebraic geometry, \\functional analysis, number theory, \\complex variables, applied mathematics, \\general mathematics, logic, \\optimization and control, statistics, \\probability, differential geometry, \\combinatorics, operator algebras, \\representation theory, classical analysis, \\dynamical systems, group theory, \\quantum algebra, rings and algebras, \\symplectic geometry, algebraic topology, \\commutative algebra, geometric topology, \\metric geometry}\\
\cline{2-3} 
& Physics (10) & \makecell{optics, fluid dynamics, atomic physics,\\ instrumentation and detectors, \\accelerator physics, general physics,\\ plasma physics, chemical physics, \\sociophysics, classical physics} \\
\cline{2-3} 
& CS (18) & \makecell{computer vision, game theory, \\information theory, machine learning, \\distributed computing, cryptography, \\ networking and internet architecture, \\computational linguistics, \\computational complexity, \\software engineering, \\artificial intelligence, systems and control, \\logic in computer science, \\cryptography and security, \\data structures and algorithms, \\programming languages, \\other computer science, databases} \\
\midrule
SDPRA 2021 & Computer Science (7) & \makecell{logic in computer science, \\distributed computing, \\software engineering, \\data structures and algorithms, \\computational linguistics, \\networking and internet architecture, \\cryptography}\\
\midrule
S2ORC & \makecell{engineering, chemistry, \\computer science, business,\\ political science, \\environmental science, physics, \\economics, geography,\\ medicine, psychology, art, \\materials science, mathematics, \\sociology, geology, \\philosophy, biology, history } & - \\
\midrule
Emerging NLP & Natural Language Processing (21) & \makecell{sign language and fingerspelling recognition,\\ rule-based machine translation (RBMT), \\transformer models, prompt engineering\\ recurrent neural networks (RNNs), \\large language models (LLMs), \\bilingual lexicon induction (BLI),\\ hate and offensive speech detection, \\email spam and phishing detection, \\fake news detection, \\fake review detection, \\aspect-based sentiment analysis (ABSA), \\dialogue state tracking (DST), \\visual question answering (VQA), \\open-domain question answering, \\multiple choice question answering (MCQA), \\nlp for for social media, \\nlp for the legal domain, \\acronyms and abbreviations detection and expansion, \\paraphrase and rephrase generation, \\named entity recognition for nested entities}\\
\bottomrule[1.5pt]
\end{tabular}}
\caption{
Detailed topic categories of four datasets. Note we classify sub-categories for arXiv, SRPRA 2021, and Emerging NLP datasets.
}
\label{tab8:data-categories}
\end{table*}

\end{document}